\documentclass{article}

\usepackage[preprint]{neurips_2026}

\usepackage[utf8]{inputenc}
\usepackage[T1]{fontenc}
\usepackage{hyperref}
\usepackage{url}
\usepackage{booktabs}
\usepackage{amsfonts}
\usepackage{amsmath}
\usepackage{amssymb}
\usepackage{nicefrac}
\usepackage{microtype}
\usepackage{xcolor}
\usepackage{graphicx}
\usepackage{array}
\usepackage{longtable}
\usepackage{caption}
\usepackage{float}
\usepackage{calc}
\usepackage{tikz}
\usetikzlibrary{positioning, arrows.meta, fit, backgrounds, calc}

\graphicspath{{figures/}}

\newcommand{\Rlin}{R^2_{\mathrm{lin}}}
\newcommand{\Rpoly}{R^2_{\mathrm{poly}}}
\newcommand{\dppl}{\Delta\mathrm{PPL}}

\title{How Linear Is a Transformer Feed-Forward Block?\\
Per-Block Linear Recoverability Is Learned, Not Architectural}

\author{%
  Stuart Whipp\\
  Independent Research\\
  \texttt{swhipp87@gmail.com}
}

\begin{document}

\maketitle

\begin{abstract}
Transformer feed-forward networks (FFNs) are often treated as nonlinear stores of model
computation, but the extent to which a trained FFN block is actually nonlinear is rarely
measured directly. We study each FFN as a position-wise map from its input activations to
its output activations, and decompose this map into the exact least-squares linear
approximation plus a residual. The held-out variance explained by this closed-form linear
map defines a block's \textbf{linear recoverability} ($\Rlin$), an optimiser-free
measurement of how much of the FFN's behaviour is captured by a single affine layer.

Across all twelve blocks of GPT-2, Pythia-160m, and llama-160m, linear recoverability is
highly heterogeneous and non-monotone across depth: adjacent blocks can range from nearly
perfectly linear ($\Rlin>0.99$) to strongly nonlinear ($\Rlin<0.3$). It is also not
predicted by activation function. GPT-2 and Pythia-160m are same-width GELU models, yet
exhibit sharply different recoverability profiles, showing that linear recoverability is a
learned property of individual trained blocks rather than an architectural property. We
then ask whether the residual is low-order multiplicative by fitting a low-rank bilinear
probe on top of the optimal linear map. The probe recovers only a few points of $R^2$, and
its gain is uncorrelated with residual nonlinearity, indicating that the unrecovered
computation is not a single position-wise product but higher-order or distributed structure.

Finally, the same measurement doubles as a targeted compression signal. Recoverable
blocks admit large single-layer replacements (GPT-2's early FFN at $\times 8$ fewer
parameters for $+0.77$ perplexity), while low-recoverability blocks flag where this is
unsafe and it exposes a methodological pitfall: trained linear baselines can badly
under-converge on ill-conditioned transformer activations, so we report the exact
closed-form least-squares ceiling throughout.
\end{abstract}

\section{Introduction}
\label{sec:intro}

Transformer FFNs are interpreted as key--value memories and as a primary store of a model's
learned computation. They are also expensive: for width $d$, the standard $d\!\to\!4d\!\to\!d$
FFN is $\approx 8d^2$ multiply-accumulates per token. Both for compression and for
interpretability, a basic question is how much of that computation is \emph{additive}
(captured by a single linear map) versus \emph{genuinely nonlinear} (requiring products of
input features). Rather than argue about it, we \textbf{measure} it.

We treat each FFN block as a position-wise map $g:\mathbb{R}^d\to\mathbb{R}^d$ (input $x$ to
output $y$) and decompose it as
\begin{equation}
  g(x) = W^{*}x + b^{*} + \rho(x),
\end{equation}
where $W^{*}x + b^{*}$ is the \textbf{best least-squares linear approximation} of $g$ over
the block's own activation distribution and $\rho$ is the residual. The fraction of output
variance the linear part explains on held-out activations is the block's \textbf{linear
recoverability} $\Rlin$; \textbf{residual nonlinearity} is $1-\Rlin$. Crucially the linear
term can be solved in \emph{closed form} (exact least squares), so $\Rlin$ is an exact,
optimiser-free, reproducible property of the block i.e. a measurement instrument rather than
a trained approximation.

We then ask what the residual \emph{is}. We fit an explicit low-rank bilinear layer
(``poly'': a linear term plus a rank-$r$ degree-2 interaction) on top of the optimal linear
map and read its gain over the ceiling as how much of the residual is \emph{low-order
multiplicative}. Multiplication here is a \textbf{probe of the residual}, not the headline.

Applying this to all twelve blocks of three pretrained decoders yields a sharp picture of
FFN structure (and a cautionary tale about how to measure it).

\paragraph{Contributions.}
\begin{enumerate}
\item \textbf{An exact per-block measure of FFN linear recoverability.} We introduce an
  activation-space distillation protocol that treats each FFN as a position-wise map and
  computes its best affine approximation by closed-form least squares. The resulting held-out
  $R^2$, $\Rlin$, is an optimiser-free measurement of how much of a trained FFN block is
  linearly recoverable.
\item \textbf{A depth survey showing that recoverability is learned, not architectural.}
  Across all twelve blocks of GPT-2, Pythia-160m, and llama-160m, $\Rlin$ is jagged and
  non-monotone across depth. Same-size GELU models disagree sharply on which blocks are
  linear, showing that recoverability is a property of the trained model and block rather
  than of the activation function or FFN architecture.
\item \textbf{A negative result on low-order multiplicative residuals.} After removing the
  exact linear component, we probe the residual with a low-rank bilinear layer. Its gain over
  the linear ceiling is small and uncorrelated with residual nonlinearity, indicating that
  the unrecovered FFN computation is not well explained as a single low-order multiplicative
  interaction.
\item \textbf{Structural and practical consequences for compression.} We show that high
  $\Rlin$ can arise either from low-rank, outlier-concentrated linear structure or from
  broadly linear high-rank structure, using reduced-rank regression and per-feature $R^2$ to
  distinguish the two (Section~\ref{sec:kinds}). The same measurement also provides a
  selective compression criterion: linearly recoverable blocks can be replaced by much
  smaller position-wise layers, while low-recoverability blocks flag themselves as poor
  candidates for such replacement.
\end{enumerate}

This also motivates the use of a closed-form baseline. Transformer activations are
ill-conditioned, so trained linear baselines can substantially understate linear
recoverability unless they are optimised to convergence. We therefore report the exact
least-squares ceiling throughout, and treat it as the measurement instrument against which
trained probes are compared.

\section{Background and Related Work}
\label{sec:background}

\begin{itemize}
\item \textbf{FFNs as key--value memory} \citep{geva2021kv} motivates asking what kind of computation FFNs perform and how compressible it is. Here, we measure
  per block.
\item \textbf{Conditioning / outlier features.} Transformer activations carry
  large-magnitude outlier features \citep{dettmers2022llmint8} that make their covariance
  ill-conditioned; this is what under-converges a naively-trained linear baseline and
  motivates the exact closed-form baseline that makes linear recoverability a reproducible
  measurement (Section~\ref{sec:survey}).
\item \textbf{Knowledge / activation distillation and probing.} We distil a sub-module's I/O
  map rather than logits; related in spirit to knowledge distillation
  \citep{hinton2015distilling}, layer-wise distillation, and model ``stitching''
  \citep{bansal2021stitching}. Measuring how much of an activation map is \emph{linearly}
  recoverable is also in the spirit of linear probing \citep{alain2017probes}, though we probe
  the FFN's own output map rather than a downstream label. Our contribution is to use
  \emph{exact} linear distillation as a measurement of structure, not only as a compression
  method.
\item \textbf{Low-rank and bilinear / polynomial layers.} The poly probe is a low-rank bilinear
  (degree-2 polynomial) layer \citep{rendle2010fm}; related to factorized bilinear pooling, deep
  polynomial networks \citep{chrysos2022pnets}, and gated linear units. We use it to interrogate
  the residual, not to win a compression race.
\item \textbf{Higher-order / multiplicative units.} Product units and sigma-pi neurons
  \citep{rumelhart1986pdp, shin1991pisigma, jayakumar2020multiplicative} compute
  multiplicative interactions; the poly probe is one such form. (A single log-space
  geometric-product layer we also tried proved numerically unstable on these targets and is
  deferred to future work, Section~\ref{sec:future}.)
\item \textbf{Gated FFNs (GLU/SwiGLU).} LLaMA-style FFNs
  $(xW_{\mathrm{gate}}\odot \mathrm{act}(xW_{\mathrm{up}}))W_{\mathrm{down}}$
  \citep{shazeer2020glu} are \emph{explicitly} multiplicative; we include one (llama-160m). A
  central finding is that this multiplicative form does \emph{not} make the residual
  recoverable by a single low-order product (Section~\ref{sec:residual}).
\item \textbf{Relation to concurrent work (this author).} A companion study
  \citep{whipp2026pi} examines a multiplicative \emph{hypernetwork} whose recruitment gate
  diagnoses multiplicative structure in a weight-generation map; there the target \emph{did}
  benefit from a product layer, in direct contrast to the FFN-residual targets here. Taken together, the two results sharpen the same point: the multiplicative benefit is
  target-specific, not generic. (This paper is self-contained; the comparison is context, not a dependency.)
\end{itemize}

\section{Method}
\label{sec:method}

\begin{figure}[t]
  \centering
  \resizebox{\linewidth}{!}{%
  \begin{tikzpicture}[
    >=Latex,
    font=\small,
    node distance=7mm and 10mm,
    box/.style={
      draw,
      rounded corners,
      align=center,
      inner sep=5pt,
      minimum height=9mm
    },
    data/.style={box, fill=blue!6, text width=24mm},
    model/.style={box, fill=gray!8, text width=28mm, minimum height=22mm},
    fitbox/.style={box, fill=green!8, text width=38mm},
    metric/.style={box, fill=red!6, text width=36mm},
    note/.style={font=\footnotesize\itshape, align=center},
    stage/.style={font=\bfseries\footnotesize, text=black!70},
    arr/.style={->, thick},
    darr/.style={->, thick, dashed},
    group/.style={
      draw=black!25,
      rounded corners,
      inner sep=6pt,
      fill=black!1
    }
  ]

  \node[data] (corpus) {Corpus\\\footnotesize WikiText / prose / math};

  \node[model, right=of corpus] (lm) {Frozen LM\\[1mm]
    \begin{tikzpicture}[baseline=-0.5ex]
      \node[draw, rounded corners, fill=orange!25,
            inner sep=3pt, font=\footnotesize] {FFN block $\ell$};
    \end{tikzpicture}
  };

  \node[data, right=of lm] (pairs) {Cached pairs\\[1mm]$(x,y)$};

  \draw[arr] (corpus) -- (lm);
  \draw[arr] (lm) -- node[above, note] {hook} (pairs);

  \begin{scope}[on background layer]
    \node[group, fit=(corpus)(lm)(pairs)] (g1) {};
  \end{scope}
  \node[stage, above=3mm of g1.north] {1. Capture FFN input--output pairs};

  \node[fitbox, right=18mm of pairs, yshift=12mm] (linear)
    {\textbf{Closed-form linear ceiling}\\[1mm]
     $W^{*}x+b^{*}$\\
     \footnotesize exact least squares};

  \node[fitbox, below=of linear] (residual)
    {\textbf{Residual probe}\\[1mm]
     $\rho(x)=g(x)-(W^{*}x+b^{*})$\\
     \footnotesize low-rank bilinear};

  \node[fitbox, below=of residual] (dense)
    {\textbf{Depth control}\\[1mm]
     Linear $\to$ GELU $\to$ Linear\\
     \footnotesize same-order budget};

  \draw[arr] (pairs.east) to[out=15,in=180] (linear.west);
  \draw[arr] (pairs.east) -- (residual.west);
  \draw[arr] (pairs.east) to[out=-15,in=180] (dense.west);

  \begin{scope}[on background layer]
    \node[group, fit=(linear)(residual)(dense)] (g2) {};
  \end{scope}
  \node[stage, above=3mm of g2.north] {2. Fit simple position-wise replacements};

  \node[metric, right=18mm of residual, yshift=8mm] (fitmetrics)
    {\textbf{Fit metrics}\\[1mm]
     \footnotesize $\Rlin$, residual gain,\\
     \footnotesize eff.\ rank, per-feat.\ $R^2$};

  \node[metric, below=of fitmetrics] (ppl)
    {\textbf{Swap metrics}\\[1mm]
     \footnotesize swap into live LM:\\
     \footnotesize $\Delta$PPL};

  \draw[arr] (linear.east) to[out=0,in=170] (fitmetrics.west);
  \draw[arr] (residual.east) -- (fitmetrics.west);
  \draw[arr] (dense.east) to[out=0,in=190] (fitmetrics.west);

  \draw[darr] (linear.east) to[out=-25,in=170]
    node[below, note, pos=0.55] {re-insert}
    (ppl.west);
  \draw[darr] (residual.east) to[out=-10,in=180] (ppl.west);
  \draw[darr] (dense.east) to[out=-5,in=190] (ppl.west);

  \begin{scope}[on background layer]
    \node[group, fit=(fitmetrics)(ppl)] (g3) {};
  \end{scope}
  \node[stage, above=3mm of g3.north] {3. Measure structure and compression cost};

  \end{tikzpicture}%
  }
  \caption{Distillation as measurement. A frozen language model is run once over a corpus, and a
  forward hook caches FFN input--output activation pairs $(x,y)$ --- one row per token for a
  chosen block. We then fit simple position-wise replacements: an exact least-squares affine map
  giving the linear ceiling, a low-rank bilinear probe of the residual, and a two-layer dense
  control. The fitted maps are evaluated both in activation space (recoverability $\Rlin$, residual
  gain, effective rank, per-feature $R^2$) and by re-insertion into the live model to measure the
  downstream perplexity change ($\Delta$PPL).}
  \label{fig:method}
\end{figure}
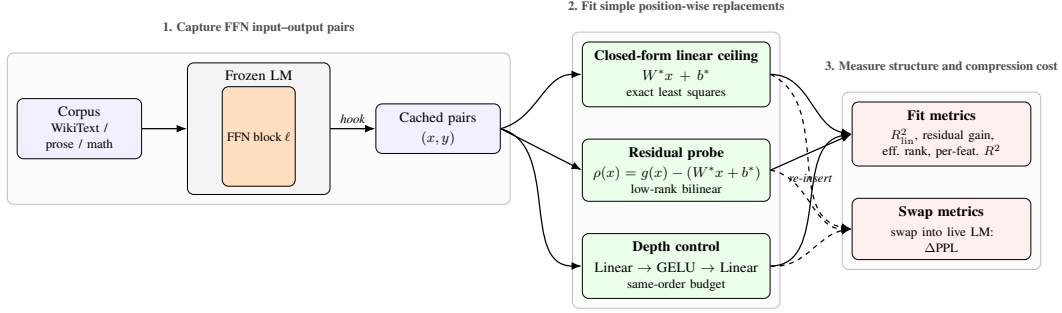

\subsection{Activation-space distillation}
\label{sec:distill}

For a chosen block we run a corpus through the frozen model under \texttt{no\_grad} and tap
the block's \texttt{.mlp} submodule with a forward hook, caching one (input
$x\in\mathbb{R}^d$, output $y\in\mathbb{R}^d$) row per token. We then fit a single
position-wise layer $f$ to minimise $\lVert f(x)-y\rVert^2$ on a held-out split. The
transformer is never back-propagated through; all learning happens on the small standalone
layer. Figure~\ref{fig:method} summarises the full pipeline.

\subsection{The linear/residual decomposition and the candidates}
\label{sec:candidates}

We instantiate the decomposition $g(x) = W^{*}x + b^{*} + \rho(x)$ with three single-layer
candidates at width $d = 768$, matched on parameter budget:
\begin{itemize}
\item \textbf{linear (closed-form)} --- \texttt{nn.Linear(d, d)}, solved in \textbf{closed
  form} by least squares: the exact best linear approximation $W^{*}x + b^{*}$, i.e.\ the
  \textbf{linear ceiling}. This is the measurement instrument, not a trained model.
  $\approx d^2$ params/MACs.
\item \textbf{poly} --- \texttt{PolyLinear}: a linear term plus a rank-$r$ bilinear (degree-2)
  term ($r = 16 \ll d$), $\approx d^2 + 2dr$ params. Used as a \textbf{probe of the residual}:
  how much of $\rho$ is low-order multiplicative. (For the residual-gain analysis of
  Section~\ref{sec:residual} we freeze its linear branch at the closed-form optimum and train
  only the bilinear branch with held-out early stopping, so its gain over the ceiling is
  $\ge 0$ by construction and isolates exactly what a low-rank product adds.)
\item \textbf{dense ($2\times$)} --- \texttt{Linear $\to$ GELU $\to$ Linear} bottleneck,
  $\approx 2d^2$ params: a depth control --- the same budget spent on an extra additive
  hidden layer instead of a product term.
\end{itemize}
The original FFN is $\approx 8d^2$ params (4.72\,M for $d = 768$), so the single-layer
candidates are 4--8$\times$ compressions. Compression is both the lens (a high-fidelity
single layer \emph{is} the evidence the block is structurally simple) and a practical payoff
in its own right (Section~\ref{sec:discussion}): on a recoverable block the single linear
layer is a near-lossless $\times 8$ parameter cut. \emph{(A single log-space
geometric-product ``Sigma-Pi'' layer was also tried and excluded: it was numerically
unstable and never beat the linear ceiling on these FFN-residual targets --- see
Section~\ref{sec:future}.)}

\subsection{Metrics}
\label{sec:metrics}

\begin{itemize}
\item \textbf{Linear recoverability} $\Rlin$ --- held-out $R^2$ of the closed-form linear map
  (variance explained); \textbf{residual nonlinearity} $= 1-\Rlin$. Reported per block, exact.
\item \textbf{Residual recovery} --- $\Rpoly - \Rlin$, the held-out $R^2$ a low-rank
  bilinear adds on top of the optimal linear map (the residual-gain probe,
  Section~\ref{sec:residual}).
\item \textbf{Effective rank} (Section~\ref{sec:kinds}) --- by \emph{reduced-rank regression}:
  the smallest $k$ whose rank-$k$ least-squares map reaches 90\% of the full closed-form $R^2$.
  Measures how many directions the linear map uses (the raw weight spectrum is uninformative
  --- outlier-scale dominated --- so we use RRR, not an SVD of $W^{*}$).
\item \textbf{Per-feature $R^2$} (Section~\ref{sec:kinds}) --- median over the $d$ output
  features of the closed-form fit; an unweighted companion to the variance-weighted $\Rlin$
  (which a few high-variance outlier features can flatter).
\item \textbf{Activation fit (worked examples)} --- mean per-row \textbf{cosine} and
  \textbf{RMSE} (raw units, comparable within a block) alongside $R^2$, in the two-block
  detail tables (Section~\ref{sec:worked}).
\item \textbf{Conjunction index} (occlusion AND-signature): how multiplicatively the response
  collapses under occlusion of two disjoint feature halves --- $\approx 0$ for an additive map.
\item \textbf{Recruitment gate} $\exp(\texttt{quad\_scale})$ and its drift over training
  (poly only) --- how much multiplicative branch the fit actually recruits.
\item \textbf{End-to-end perplexity.} We re-insert each fitted layer into the live model and
  measure WikiText-2 perplexity: (a)~\textbf{zero-shot $\dppl$} (swap, nothing else changed);
  (b)~\textbf{healed $\dppl$} after fine-tuning \emph{only the swapped layer} for a small
  budget; and (c)~a \textbf{heal-original baseline} --- the original FFN given the \emph{same}
  heal budget --- so healed numbers are read against an equally-adapted original (see
  Section~\ref{sec:ppl}).
\end{itemize}

\subsection{Compute / complexity}
\label{sec:compute}

Per token the original FFN costs $\approx 8d^2$ MACs (the $4\times$ inner expansion). The
single-layer replacements collapse this: \textbf{linear} and \textbf{poly} are $\approx d^2$
(poly $= d^2 + 2dr \approx 1.04 d^2$ at $r = 16$), an $\approx 8\times$ FLOP reduction;
\textbf{dense ($2\times$)} is $\approx 2d^2$. All are $\mathcal{O}(d^2)$ versus the FFN's
$\mathcal{O}(d^2)$ with an $8\times$ larger constant. The poly probe adds only the small
rank-$r$ bilinear term over a plain linear map, so wherever a linear map already suffices it
is nearly free.

\section{Experimental Setup}
\label{sec:setup}

\begin{itemize}
\item \textbf{Models:} three pretrained decoders at matched width/depth ($d = 768$, 12
  blocks): \textbf{GPT-2} \citep{radford2019gpt2} (124\,M, GELU FFN), \textbf{Pythia-160m}
  \citep{biderman2023pythia} (GPT-NeoX, GELU FFN) --- a second, independently-trained GELU
  model that lets us test whether linearity is a GELU property --- and \textbf{llama-160m}
  (\texttt{JackFram/llama-160m} on the Hub; LLaMA architecture \citep{touvron2023llama}, SiLU
  \textbf{SwiGLU} FFN, intermediate 3072) for a different FFN family. A
  \textbf{TinyLlama-1.1B} SwiGLU model is reported as a \emph{scale} probe (see below) not as a fourth survey datapoint.
\item \textbf{Blocks:} all twelve blocks for the depth survey (Sections~\ref{sec:survey} --%
 \ref{sec:kinds}); an \emph{early} block (index 1) and a \emph{deep} block (index 10) for the
 fidelity and perplexity tables from the worked sample (Sections~\ref{sec:worked} -- \ref{sec:ppl}).
  Blocks are zero-indexed throughout: ``early~(1)'' is the second transformer block and
  ``deep~(10)'' the eleventh.
\item \textbf{Corpus:} WikiText-2-raw \citep{merity2017pointer}. ${\sim}15$--$30$\,k token
  rows captured per block (seq len 128); held-out 20\% for fit metrics; perplexity on the
  test split. None of the models were trained on Wikipedia, so absolute PPL is higher than
  each model's headline number. We report \emph{deltas} against each model's own fixed base.
\item \textbf{Fit:} the linear baseline is solved in \textbf{closed form} (exact least
  squares, the linear ceiling); the poly / 2-layer candidates are trained with AdamW
  (converged, verified against the ceiling). For the Section~\ref{sec:residual} residual probe
  poly's linear branch is frozen at the closed-form optimum and only its bilinear branch is
  trained, with held-out early stopping.
\item \textbf{Heal:} 200 steps, lr 1e-4, swapped-layer-only (and original-only for the
  per-block baseline).
\item \textbf{Seeds:} 42 / 43 / 44; stochastic fits report mean $\pm$ std. The closed-form
  linear map is deterministic and reported as a single number. \emph{Note on what the
  $\pm$std captures:} the cached activations $(X, Y)$ are identical across seeds (a
  \texttt{no\_grad} eval-mode forward pass is deterministic) and the train/val split is a
  fixed tail, so a seed varies only layer initialisation and minibatch order. The very small
  spreads (typically $\pm 0.000$--$0.001$ $R^2$) therefore measure \emph{optimisation
  reproducibility}, not corpus- or split-sampling uncertainty. They say the converged fits
  are stable, not that the numbers are immune to a change of data. We probe the larger and more
 honest perturbations directly in Section~\ref{sec:robust}: a blocked $k$-fold
  cross-validation that gives the ceiling a real data-split CI (mean std $0.024$ $R^2$), and
  changes of corpus \emph{domain}.
\end{itemize}

\section{Results}
\label{sec:results}

\subsection{Linear recoverability across depth and models}
\label{sec:survey}

\begin{figure}[t]
  \centering
  \includegraphics[width=\linewidth]{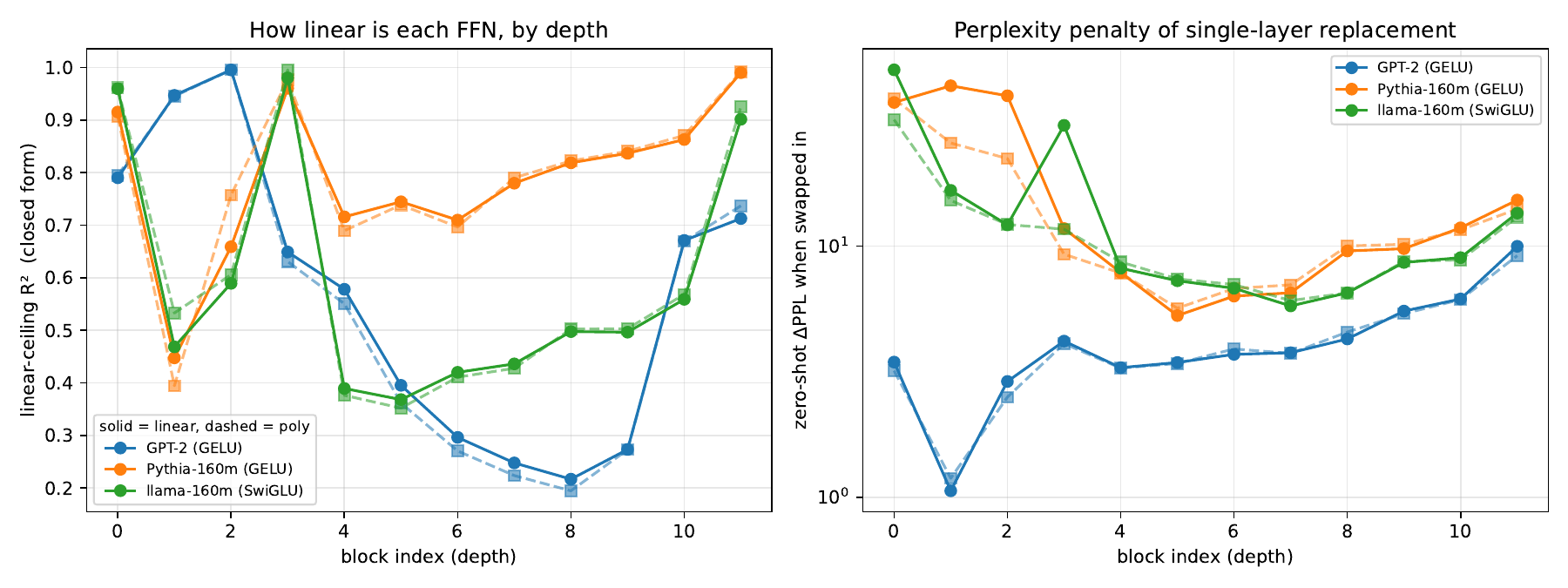}
  \caption{Per-block linear recoverability $\Rlin$ (exact closed-form ceiling) and zero-shot
  swap $\dppl$ across all twelve blocks of GPT-2, Pythia-160m, and llama-160m. Linearity is
  jagged and non-monotone, and the profiles differ sharply across models, including the
  two same-size GELU models (GPT-2, Pythia).}
  \label{fig:depth}
\end{figure}

Our central measurement is the per-block linear ceiling $\Rlin$ across all twelve blocks of
all three models (Figure~\ref{fig:depth}). Two things stand out immediately.

\paragraph{Linearity is jagged and non-monotone across depth, and model-specific.} GPT-2's
recoverability by block runs $0.79, 0.95, \mathbf{0.996}, 0.65, 0.58, 0.40, 0.30,
\mathbf{0.25}, 0.22, 0.27, 0.67, 0.71$ --- near-perfectly linear at block 2, strongly
\emph{nonlinear} through the middle (blocks 5--9 sit at ${\sim}0.2$--$0.4$), then partially
linear again at the end. Pythia runs $0.92, 0.45, 0.66, \mathbf{0.96}, 0.72, 0.74, 0.71,
0.78, 0.82, 0.84, 0.86, \mathbf{0.99}$ rising toward the deep end. llama runs $0.96, 0.47,
0.59, \mathbf{0.98}, 0.39, 0.37, 0.42, 0.44, 0.50, 0.50, 0.56, 0.90$ --- linear at the ends,
nonlinear in the middle. There is no universal ``early-nonlinear / deep-linear'' rule; each
model has its own jagged profile, and individual blocks can be almost perfectly linear right
next to strongly nonlinear ones (GPT-2 block 2 at 0.996, block 8 at 0.22).

\paragraph{A cautionary methodological note (why the closed-form baseline).} A \emph{trained}
linear baseline can badly understate recoverability on these activations. For example, after
3\,000 SGD steps the GPT-2 early block reaches only $R^2 \approx 0.25$, even though the exact
closed-form map scores 0.95 on the same block. The gap is an \textbf{optimisation artifact},
not nonlinearity: GPT-2's FFN activations are severely ill-conditioned, with an input
covariance condition number $\approx 3\times10^7$ and one output feature carrying
${\sim}100\times$ the median variance (the transformer outlier-feature phenomenon,
\citealp{dettmers2022llmint8}). A plain linear layer therefore needs ${\sim}15\times$ more steps
(${\sim}50\,000$) to converge, whereas the factorized poly layer self-conditions and reaches the
ceiling by ${\sim}3\,000$. Every $R^2$ and $\dppl$ in this paper therefore uses the closed-form
linear ceiling, which is exact and optimiser-independent.

\subsection{Linearity is learned, not architectural}
\label{sec:learned}

Is linear recoverability set by the activation function? It is not. \textbf{GPT-2 and
Pythia-160m are the same size ($d = 768$, 12 blocks) with the same GELU activation, yet have
opposite depth profiles} (Table~\ref{tab:learned}).

\begin{table}[t]
  \centering
  \caption{Closed-form linear ceilings $\Rlin$ (single deterministic value); zero-shot
  linear-swap $\dppl$ in parentheses, against each model's own base. Two same-size GELU models
  (GPT-2, Pythia) disagree on which blocks are linear.}
  \label{tab:learned}
  \begin{tabular}{lccc}
    \toprule
    Block & GPT-2 (GELU) & Pythia-160m (GELU) & llama-160m (SwiGLU) \\
    \midrule
    early (1) & \textbf{0.95} ($+1.1$ PPL) & \textbf{0.45} ($+63$ PPL) & \textbf{0.47} ($+19$ PPL) \\
    deep (10) & 0.67 ($+5.5$ PPL) & 0.86 ($+12$ PPL) & 0.56 ($+8.8$ PPL) \\
    \bottomrule
  \end{tabular}
\end{table}

GPT-2's early block is 95\% linear and a linear swap is nearly free ($+1.1$ PPL); the
\emph{same-size, same-GELU} Pythia early block is only 45\% linear and a linear swap is
catastrophic ($+63$ PPL). Two same-width, same-depth GELU models disagree on which blocks
are linear and by how much, hence ``GELU FFNs are near-linear'' is false as a general claim.
\textbf{Linear recoverability is a property of the specific trained model and block, not of
its architecture or activation function.} GPT-2's
near-linear early FFN is an outlier, not a rule.
(llama's SwiGLU, despite being \emph{explicitly} multiplicative in form, is also genuinely
nonlinear in the early block. Being multiplicative in form does not make it more linearly
recoverable; Section~\ref{sec:residual} shows it does not make the residual recoverable
either.)

\subsection{Is the residual multiplicative? A low-rank bilinear probe}
\label{sec:residual}

\begin{figure}[t]
  \centering
  \includegraphics[width=0.58\linewidth]{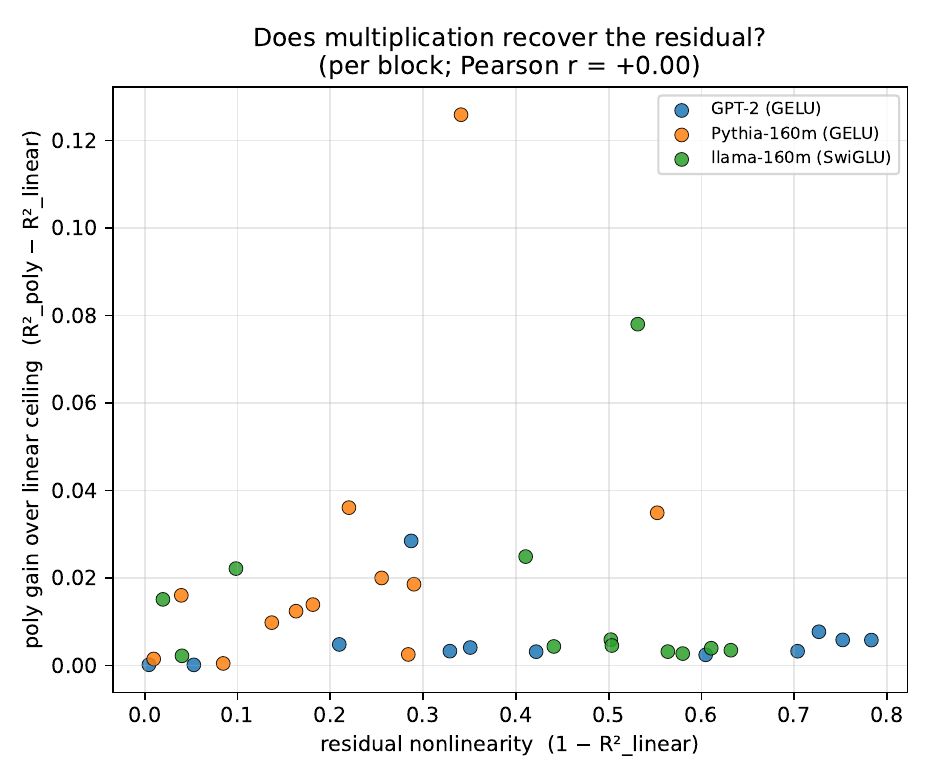}
  \caption{Residual nonlinearity ($1-\Rlin$) vs.\ the gain a rank-16 bilinear probe adds over
  the linear ceiling ($\Rpoly-\Rlin$), per block. The gain is small everywhere and
  \emph{uncorrelated} with residual nonlinearity (Pearson $r\approx 0$): the residual is not
  recovered by a low-rank degree-2 (bilinear) layer.}
  \label{fig:residual}
\end{figure}

Having measured the linear part exactly, we ask what the \emph{residual} ($1-\Rlin$) is: is it
low-order multiplicative, the kind of thing a single bilinear layer could recover? We freeze
poly's linear branch at the closed-form optimum and train only its rank-16 bilinear branch
with held-out early stopping, so the gain $\Rpoly - \Rlin$ is $\ge 0$ by construction and
measures exactly what a low-order product adds on top of the optimal linear map. Plotting this
gain against residual nonlinearity for all 36 blocks (Figure~\ref{fig:residual}) gives a clean
negative answer.

\paragraph{The residual is mostly not low-order multiplicative.} The bilinear probe recovers
only a few points of $R^2$ almost everywhere (median gain $< 0.01$; max 0.13), and \textbf{its gain does not scale with residual nonlinearity}: across all 36 blocks
the Pearson correlation between $(1-\Rlin)$ and the poly gain is $\mathbf{+0.004}$, i.e.\ none.
Residual size does not predict whether the probe helps. For instance, llama block 4 has a
large residual (0.61) but the probe recovers almost nothing ($+0.004$), whereas llama block 1
has a \emph{smaller} residual (0.53) yet gains $+0.078$; GPT-2's most nonlinear blocks
(residual $> 0.7$) likewise gain $\le 0.008$. The few blocks where the probe helps at all ---
Pythia block 2 ($+0.126$), llama block 1 ($+0.078$), and Pythia block 7 ($+0.036$) --- are
idiosyncratic, not the most nonlinear. Within Pythia alone there is a mild positive trend ($r = +0.46$, driven by blocks 1--2),
but it does not hold for GPT-2 ($+0.05$) or llama ($-0.01$). \textbf{So the residual is not
captured by this low-rank degree-2 probe --- a single position-wise product does not recover
it and ``the FFN is multiplicative in form'' (SwiGLU's gate) neither makes the block more
linearly recoverable nor its residual more bilinearly recoverable.} This indicates that
multiplicative form does not predict multiplicative recoverability; we interpret the
unrecovered part as higher-order or distributed structure in the Discussion
(Section~\ref{sec:discussion}), not as something this probe pins down.

Consistent diagnostics: the closed-form linear map reads conjunction index 0.000 (purely
additive, as it must); the 2-layer GELU bottleneck reads high ($\approx 0.90$ --- genuinely
conjunctive); poly sits in between ($\approx 0.31$). And on a near-linear target poly's
multiplicative-recruitment gate \emph{de-recruits} during fitting ($\exp(\texttt{quad\_scale})$
$0.135 \to 0.055$) the optimiser actively shrinks the quadratic branch when there is little
multiplicative structure to recruit, agreeing with the near-zero gains above.

\subsection{What kind of linearity? Effective rank and per-feature recoverability}
\label{sec:kinds}

\begin{figure}[t]
  \centering
  \includegraphics[width=0.72\linewidth]{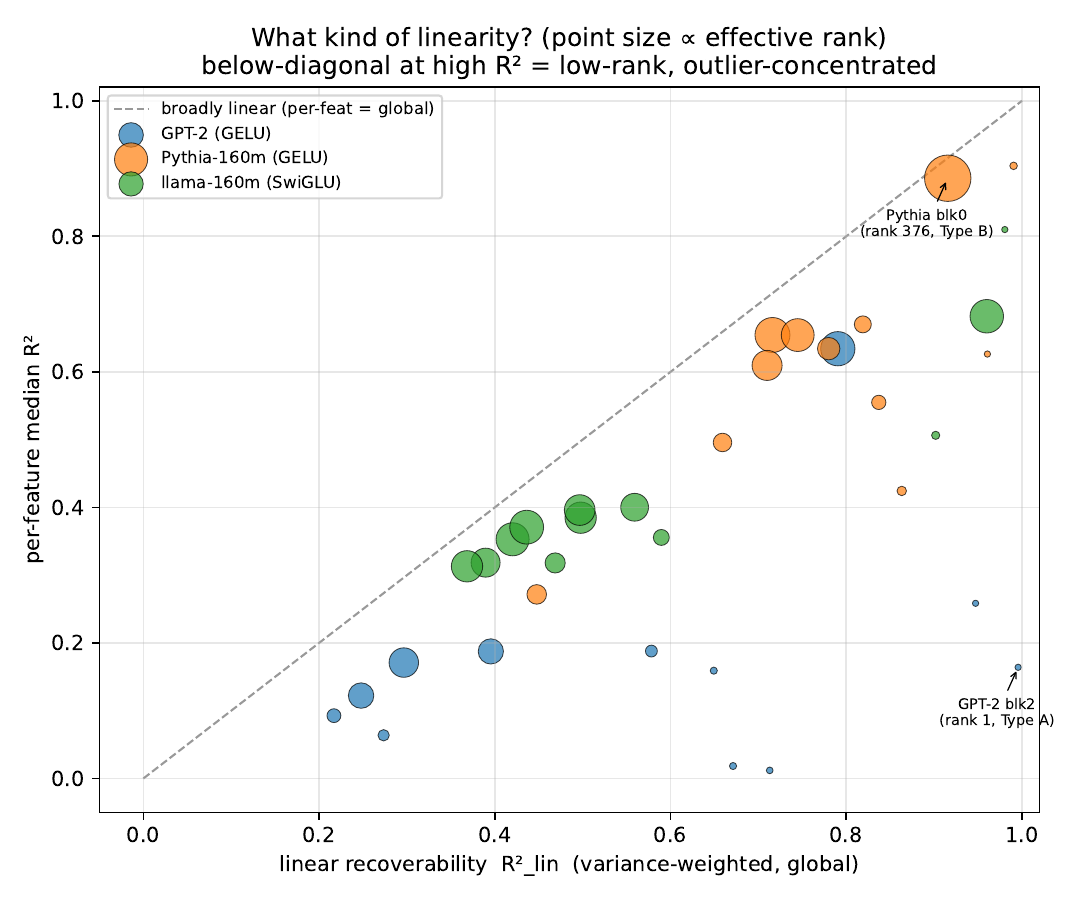}
  \caption{Global $\Rlin$ vs.\ per-feature median $R^2$, point size $\propto$ effective rank.
  Points near the diagonal are \emph{broadly linear} (high rank); points far below the diagonal
  at high $\Rlin$ are \emph{low-rank, outlier-concentrated} (rank $\approx 1$). A high $\Rlin$
  hides two structurally opposite regimes.}
  \label{fig:kinds}
\end{figure}

Linear recoverability $\Rlin$ is variance-weighted, so a high value can arise two very
different ways. To tell them apart we add two scale-aware structural readouts of the same
closed-form fit: the \textbf{effective rank} of the linear map by \emph{reduced-rank
regression} (the smallest $k$ whose rank-$k$ least-squares map reaches 90\% of the full
closed-form $R^2$, used as a proxy for how many directions the linear map uses,
since the raw weight spectrum is dominated by outlier-feature scale, Section~\ref{sec:discussion})
and the \textbf{per-feature $R^2$} (median over the $d$ output features, unweighted by
variance). Figure~\ref{fig:kinds} plots the two against each other for all 36 blocks.

\paragraph{A high $\Rlin$ hides two structurally opposite regimes.}
\begin{itemize}
\item \emph{Low-rank, outlier-concentrated} (effective rank $\approx 1$--$6$, per-feature $R^2$
  low). GPT-2 block 2 has $\Rlin = 0.996$ but a per-feature median $R^2$ of only \textbf{0.16}
  and an effective rank of \textbf{1} .Almost all of its \emph{variance-weighted}
  recoverability lives in a single dominant output direction, and the median individual feature
  is barely linear. The same pattern holds for GPT-2 block 1 ($0.95 / 0.26 / \text{rank }1$),
  Pythia block 3 ($0.96 / 0.63 / \text{rank }1$) and block 11 ($0.99 / 0.90 / \text{rank }4$),
  and llama blocks 3 and 11.
\item \emph{High-rank, broadly linear} (effective rank $\approx 190$--$380$, per-feature $R^2$
  high). Pythia block 0 has $\Rlin = 0.92$, a per-feature median of \textbf{0.89}, and an
  effective rank of \textbf{376} --- genuinely linear across most features. llama block 0
  ($0.96 / 0.68 / \text{rank }193$) and GPT-2 block 0 ($0.79 / 0.63 / \text{rank }202$) are
  likewise broadly linear.
\end{itemize}
So \textbf{$\Rlin$ and effective rank are largely decoupled}: high recoverability occurs both at
rank 1 and at rank ${\sim}380$. This \emph{reconciles} rather than undercuts
Section~\ref{sec:survey}: GPT-2 block 2's 0.996 is real in the variance-weighted,
downstream-relevant sense. A linear swap costs only $+0.77$ PPL (Section~\ref{sec:ppl})
precisely because the residual stream's high-variance direction is the one that is linear, even though most of the block's features are not individually linear. The effective rank says
what \emph{kind} of linear object the block is: a low-rank block behaves like a single (or few)
linear ``key--value'' memory direction(s) (cf.\ \citealp{geva2021kv}) and would compress further
still (to a near-rank-1 linear map), whereas a high-rank block needs the full $d^2$ budget.
The decomposition also sharpens our ``learned, not architecture'' claim (Section~\ref{sec:learned}):
Pythia block 0 (broadly linear, per-feature 0.89) and GPT-2 block 1 (outlier-linear, per-feature
0.26) have nearly the same $\Rlin$ yet opposite internal structure, so even the \emph{form} of a
block's linearity is a learned, per-block property. (Because effective rank is measured by
variance-weighted RRR it inherits $\Rlin$'s outlier-weighting; the per-feature median $R^2$ is
its unweighted complement, and we read the two together.)

\subsection{Worked example: two blocks in detail (fidelity)}
\label{sec:worked}

The depth survey (Sections~\ref{sec:survey}--\ref{sec:residual}) is the evidence; this section
zooms in on one early and one deep block of two contrasting models
(Tables~\ref{tab:fidelity-gpt2}--\ref{tab:fidelity-llama}) to show the underlying fit
quality (cosine, RMSE, parameters/compression). \textbf{These two blocks are illustrative, not
representative --- Section~\ref{sec:survey} shows recoverability is jagged across depth, so no
two-block pick is a summary of a model.} The \texttt{linear} row is the exact closed-form ceiling
(single value); poly / dense ($2\times$) / poly ($2\times$) are trained to convergence (mean
$\pm$ std over seeds 42/43/44; RMSE omitted for the poly ($2\times$) rows, which come from the
separate depth-control run). \emph{(These tables use the higher-token multi-seed runs,
${\sim}30$\,k tokens/block, so the closed-form ceilings differ slightly from the 15\,k-token
depth survey of Sections~\ref{sec:survey}--\ref{sec:learned} --- e.g.\ GPT-2 deep 0.74 here vs
0.67 in the survey; the difference is sampling of the activation distribution, not the map, and
is well within the cross-corpus robustness discussed in Section~\ref{sec:limitations}.)}

\begin{table}[t]
  \centering
  \small
  \setlength{\tabcolsep}{5pt}
  \caption{GPT-2 (GELU), original FFN 4{,}722{,}432 params.}
  \label{tab:fidelity-gpt2}
  \begin{tabular}{llrrrrr}
    \toprule
    Block & Layer & Params & Compress & $R^2$ & Cosine & RMSE \\
    \midrule
    early (1) & linear (closed-form) & 590{,}592   & $\times 8.0$ & 0.954 & 0.650 & 0.342 \\
    early (1) & poly                 & 628{,}224   & $\times 7.5$ & $0.956 \pm 0.000$ & $0.659 \pm 0.000$ & 0.335 \\
    early (1) & dense ($2\times$)    & 1{,}181{,}184 & $\times 4.0$ & $0.960 \pm 0.000$ & $0.693 \pm 0.001$ & 0.321 \\
    early (1) & poly ($2\times$)     & 1{,}256{,}448 & $\times 3.8$ & $0.961 \pm 0.000$ & $0.708 \pm 0.000$ & --- \\
    deep (10) & linear (closed-form) & 590{,}592   & $\times 8.0$ & 0.736 & 0.845 & 1.465 \\
    deep (10) & poly                 & 628{,}224   & $\times 7.5$ & $0.746 \pm 0.000$ & $0.848 \pm 0.000$ & 1.437 \\
    deep (10) & dense ($2\times$)    & 1{,}181{,}184 & $\times 4.0$ & $0.768 \pm 0.000$ & $0.864 \pm 0.000$ & 1.373 \\
    deep (10) & poly ($2\times$)     & 1{,}256{,}448 & $\times 3.8$ & $0.768 \pm 0.000$ & $0.863 \pm 0.000$ & --- \\
    \bottomrule
  \end{tabular}
\end{table}

\begin{table}[t]
  \centering
  \small
  \setlength{\tabcolsep}{5pt}
  \caption{llama-160m (SwiGLU), original FFN 7{,}077{,}888 params.}
  \label{tab:fidelity-llama}
  \begin{tabular}{llrrrrr}
    \toprule
    Block & Layer & Params & Compress & $R^2$ & Cosine & RMSE \\
    \midrule
    early (1) & linear (closed-form) & 590{,}592   & $\times 12.0$ & 0.402 & 0.588 & --- \\
    early (1) & poly                 & 628{,}224   & $\times 11.3$ & $0.475 \pm 0.001$ & $0.625 \pm 0.001$ & --- \\
    early (1) & dense ($2\times$)    & 1{,}181{,}184 & $\times 6.0$  & $0.483 \pm 0.001$ & $0.692 \pm 0.001$ & --- \\
    early (1) & poly ($2\times$)     & 1{,}256{,}448 & $\times 5.6$  & $0.483 \pm 0.002$ & $0.695 \pm 0.001$ & --- \\
    deep (10) & linear (closed-form) & 590{,}592   & $\times 12.0$ & 0.527 & 0.711 & --- \\
    deep (10) & poly                 & 628{,}224   & $\times 11.3$ & $0.539 \pm 0.001$ & $0.718 \pm 0.001$ & --- \\
    deep (10) & dense ($2\times$)    & 1{,}181{,}184 & $\times 6.0$  & $0.585 \pm 0.001$ & $0.744 \pm 0.001$ & --- \\
    deep (10) & poly ($2\times$)     & 1{,}256{,}448 & $\times 5.6$  & $0.588 \pm 0.000$ & $0.747 \pm 0.000$ & --- \\
    \bottomrule
  \end{tabular}
\end{table}

On GPT-2's near-linear early block all candidates land together at $R^2 \approx 0.95$--$0.96$:
poly adds $+0.002$, dense ($2\times$) $+0.006$ --- the residual is small and not bilinearly
recoverable (Section~\ref{sec:residual}). On llama's genuinely nonlinear early block the ceiling
is 0.40 and the spread is wider (poly $+0.073$, dense ($2\times$) $+0.081$), but even the 2-layer
additive control recovers only a fraction of the residual; consistent with the survey's
verdict that the residual is not low-order-bilinear-recoverable and only partly reachable by
any single position-wise layer, multiplicative or not.

\paragraph{Multiplicative depth $\approx$ additive depth.} The poly ($2\times$) row
(\texttt{PolyLinear $\to$ GELU $\to$ PolyLinear}, the multiplicative analog of dense ($2\times$))
lets us ask whether \emph{adding multiplicativity to a two-layer candidate} helps beyond the
depth itself. It does not: poly ($2\times$) matches dense ($2\times$) to within $\le 0.002$ $R^2$
at every block (GPT-2 early 0.961 vs 0.960, GPT-2 deep 0.768 vs 0.768, llama early 0.483 vs 0.482,
llama deep 0.588 vs 0.585) and at a slightly \emph{larger} budget, so the tiny edge is within
noise. The gain of the two-layer candidates over a single layer is therefore the \textbf{depth}
(an extra hidden nonlinearity), not the \textbf{multiplicativity}: once a hidden layer is present,
making its projections explicitly bilinear adds essentially nothing. This is the depth-axis
counterpart of Section~\ref{sec:residual}'s width-axis result. Neither multiplicative
\emph{width} (poly's bilinear term) nor multiplicative \emph{depth} (poly ($2\times$)) recovers
the FFN residual that an equal budget of plain additive capacity does not.

\paragraph{Scale probe (TinyLlama-1.1B SwiGLU, $d = 2048$).} As an external-validity check at
${\approx}9\times$ the width we probe two blocks of TinyLlama-1.1B (SwiGLU, $12\times$ larger
FFN, 34.6\,M params; base PPL 19.8) with \emph{scale-aware} fitting (the closed-form-seeded poly
of Section~\ref{sec:residual} and gradient-clipped training, which keep the candidates stable
where a naive lr-1e-3 fit diverges at this width). The fits are stable but the signal is weak: the
closed-form linear ceiling is only $R^2 \approx 0.04$--$0.07$ \emph{globally}, and the
\textbf{per-feature} $R^2$ is actually \emph{negative} (median $-0.02$ early, $-0.21$ deep) --- the
median output feature is predicted worse than its own mean, so the small global $R^2$ is propped
up by a few high-variance features rather than a broadly good fit. poly adds nothing over linear
($\le 0.001$) and dense ($2\times$) only a little (to 0.09/0.25 global), with zero-shot swaps
costing $+2.9$--$5.1$ PPL. We therefore treat TinyLlama as \emph{directional only}. It confirms
that a billion-parameter SwiGLU FFN strongly resists single-position-wise-layer distillation
(consistent with, and stronger than, llama-160m) but is not a clean survey datapoint.
Methodologically it also shows the value of the per-feature $R^2$ as a stricter companion to the
variance-weighted global $R^2$, which here is the \emph{optimistic} reading.

\subsection{Downstream perplexity: \texorpdfstring{$R^2$}{R2} and \texorpdfstring{$\dppl$}{dPPL} dissociate}
\label{sec:ppl}

Re-inserting each fitted layer into the live model and measuring WikiText-2 perplexity reveals
that \textbf{activation-fit $R^2$ and downstream perplexity impact measure different things.}
From the depth survey:
\begin{itemize}
\item \textbf{High $R^2$ does not imply low $\dppl$.} llama block 0 has $\Rlin = 0.96$ yet a
  linear swap costs \textbf{$+76$ PPL}; block 3 has $\Rlin = 0.98$ yet costs \textbf{$+40$ PPL}.
  Pythia block 0 ($R^2$ 0.92) costs $+52$ PPL. Early blocks are \textbf{perplexity-critical}
  almost regardless of how linearly fittable they are i.e. a near-perfect activation fit can still
  wreck the model because the residual stream is sensitive to small early perturbations.
\item \textbf{The criticality is model-specific.} GPT-2 block 0 ($R^2$ 0.79) costs only $+2.6$
  PPL, while llama / Pythia block 0 cost $+76$ / $+52$ --- the same depth, very different
  downstream fragility.
\item \textbf{Where the multiplicative probe helps downstream, it is on these critical blocks.}
  Although poly's \emph{$R^2$} gain is tiny and uncorrelated with residual nonlinearity
  (Section~\ref{sec:residual}), its \emph{$\dppl$} benefit is concentrated on early,
  perplexity-critical blocks: llama block 0 $+76 \to +43$, block 3 $+40 \to +12$; Pythia block 1
  $+63 \to +33$, block 2 $+56 \to +27$ --- and $\approx 0$ on blocks 4--11. Strikingly, llama
  block 0 gets a 33-point PPL reduction from poly while its $R^2$ barely moves ($+0.002$),
  underlining that $R^2$ is a poor proxy for downstream impact and that both should be reported.
\end{itemize}

The two-block detail (Section~\ref{sec:worked} models, base PPL GPT-2 64.25 / llama 41.17)
confirms the fidelity ordering downstream (Table~\ref{tab:ppl}).

\begin{table}[t]
  \centering
  \small
  \caption{Zero-shot and healed $\dppl$ (lower is better). \emph{orig (healed)} is the original
  FFN given the same per-block heal budget as an equally-adapted baseline.}
  \label{tab:ppl}
  \begin{tabular}{lllrr}
    \toprule
    Model & Block & Layer & $\dppl$ (zero-shot) & $\dppl$ (healed) \\
    \midrule
    GPT-2 & early (1) & linear (closed-form) & $\mathbf{+0.77}$ & $-6.92$ \\
    GPT-2 & early (1) & poly                 & $\mathbf{+0.50 \pm 0.01}$ & $-11.77 \pm 0.05$ \\
    GPT-2 & early (1) & dense ($2\times$)    & $\mathbf{+0.58 \pm 0.04}$ & $-17.25 \pm 0.07$ \\
    GPT-2 & early (1) & \emph{orig (healed)} & --- & $\mathbf{-20.70 \pm 0.06}$ \\
    GPT-2 & deep (10) & linear (closed-form) & $+4.70$ & $+2.02$ \\
    GPT-2 & deep (10) & poly                 & $+4.65 \pm 0.05$ & $-5.77 \pm 0.20$ \\
    GPT-2 & deep (10) & dense ($2\times$)    & $+4.50 \pm 0.06$ & $-14.36 \pm 0.11$ \\
    GPT-2 & deep (10) & \emph{orig (healed)} & --- & $\mathbf{-20.39 \pm 0.01}$ \\
    llama & early (1) & linear (closed-form) & $+16.21$ & $-5.64$ \\
    llama & early (1) & poly                 & $+14.52 \pm 0.06$ & $-6.52 \pm 0.09$ \\
    llama & early (1) & dense ($2\times$)    & $+13.83 \pm 0.18$ & $-6.16 \pm 0.11$ \\
    llama & early (1) & \emph{orig (healed)} & --- & $\mathbf{-16.01}$ \\
    \bottomrule
  \end{tabular}
\end{table}

\paragraph{The healing confound and the heal-original control.} Healed $\dppl$ is negative for
most candidates --- a \emph{healed} single layer can score below stock GPT-2. This is not
``compression improves the model'': healing fine-tunes the swapped layer on WikiText-2
\emph{train}, but the stock models never saw Wikipedia, so the healed variant gains a sliver of
in-domain adaptation the base lacks, and more parameters means more headroom to absorb it (why the
wider dense$2\times$ heals ``best''). We control with the \textbf{heal-original baseline} --- the
\emph{same} per-block FFN given the \emph{same} heal budget, rest frozen (a per-block quantity, not
whole-model fine-tuning): $-20.70$ (GPT-2 early) and $-20.39$ (GPT-2 deep). The full 4.7\,M-param
FFN, healed identically, captures almost all the in-domain headroom; no compressed candidate closes
the gap (best, early dense ($2\times$) at $-17.25$, is ${\sim}3.5$ PPL short). So healing does not
overturn the zero-shot story --- what the small layers cannot recover is precisely the extra
capacity of the wide FFN. We treat zero-shot $\dppl$ as the fidelity metric and the
(healed $-$ heal-original) gap as a capacity probe; both agree.

\subsection{Cross-domain robustness}
\label{sec:robust}

\begin{figure}[t]
  \centering
  \includegraphics[width=0.9\linewidth]{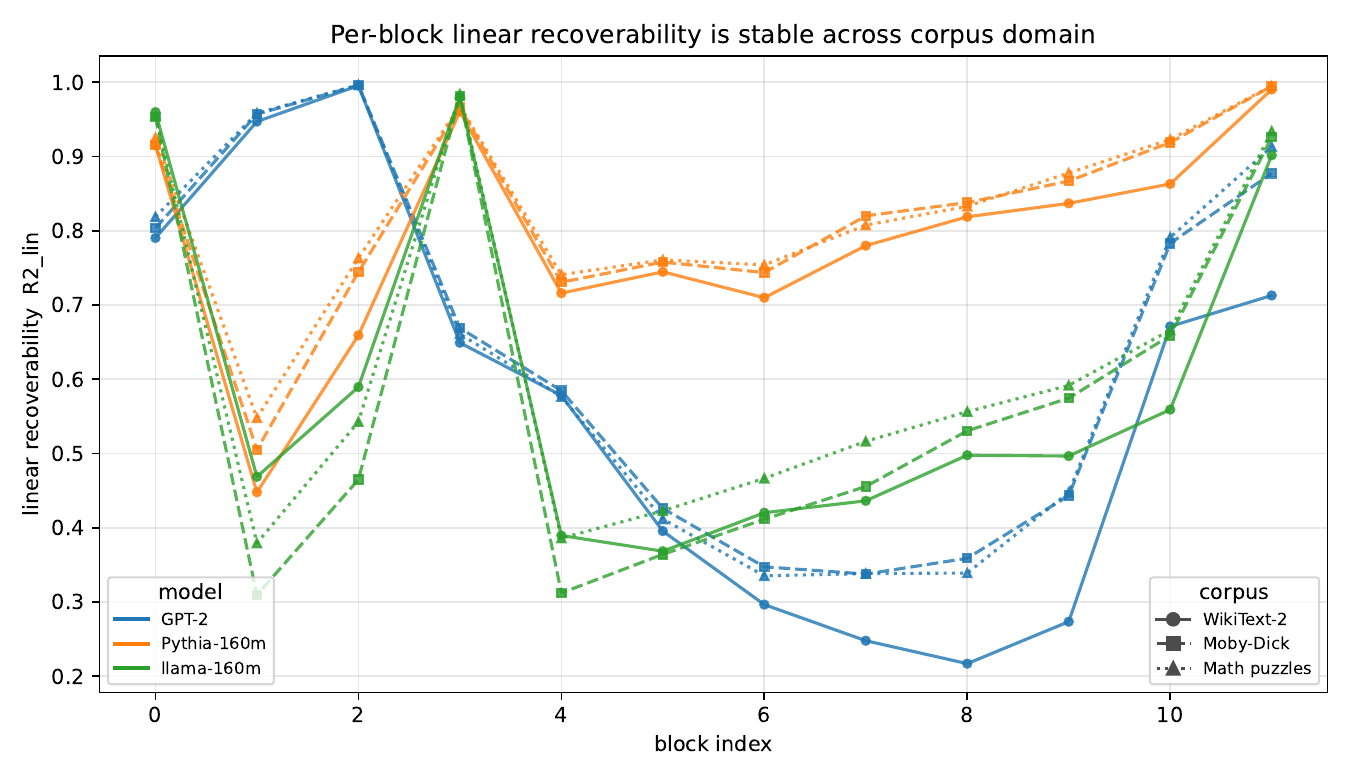}
  \caption{Per-block linear recoverability $\Rlin$ across three corpus domains (colour = model,
  linestyle = corpus). The three profiles overlay closely per model, so the recoverability profile
  is a property of the model, not the corpus.}
  \label{fig:corpus}
\end{figure}

Because $\Rlin$ is measured over the activation distribution a corpus induces, a fair robustness
check is a \emph{different-domain} corpus --- not merely a larger same-domain one (WikiText-103 is
still Wikipedia). We re-run the survey on \textbf{two} further domains: literary prose (Project
Gutenberg, \emph{Moby-Dick}) and mathematical/logical puzzles (Dudeney, \emph{Amusements in
Mathematics}), each ${\sim}1$\,MB at the same 15\,k-token budget, and compare the per-block profile
to WikiText-2 (Figure~\ref{fig:corpus}).

\paragraph{The linear-recoverability profile is a property of the model, not the corpus.} Across
all three domains the per-block ceilings track tightly: Moby-Dick vs WikiText-2 gives Pearson
$r = 0.97 / 0.99 / 0.95$ (GPT-2 / Pythia / llama) and the math-puzzle corpus $0.97 / 0.98 / 0.97$
(Spearman $\ge 0.87$ throughout); the mean absolute shift in $R^2$ is only 0.03--0.07 (worst-case
$\approx 0.20$ on a single block). Every qualitative claim survives intact on both out-of-domain
corpora: the jagged non-monotone depth profiles (Section~\ref{sec:survey}), the GPT-2-vs-Pythia
reversal that grounds ``learned structure, not defined architecture'' (Section~\ref{sec:learned}), and each model's
near-linear and strongly nonlinear blocks. Absolute ceilings move a little since
the input distribution changes, but the \emph{shape} and the cross-model \emph{contrasts} that
carry the paper's claims do not.

\paragraph{The downstream and multiplicative findings are corpus-robust too.} Repeating the full
depth sweep (closed-form ceiling + seeded-poly + zero-shot $\dppl$, with its own held-out
train/test split) on \emph{Moby-Dick} reproduces both secondary results of Section~\ref{sec:ppl}:
the \textbf{$R^2$--$\dppl$ dissociation} (llama block 0 has $\Rlin = 0.95$ yet a linear swap costs
\textbf{$+72$ PPL}; block 3 $\Rlin = 0.98$ yet $+57$ PPL; Pythia block 2 $+31$) and \textbf{poly's
$\dppl$ benefit concentrated on the early perplexity-critical blocks} (llama block 0 $+72\to+40$,
block 3 $+57\to\mathbf{+17}$; Pythia block 2 $+31\to+15$; $\approx 0$ elsewhere). So the linear
\emph{and} the multiplicative/downstream stories hold on out-of-domain text.

\paragraph{A data-split (not just seed) confidence interval.} Finally, to give the ceiling a
variance over \emph{data} rather than the deterministic seed-std of Section~\ref{sec:setup}, we run
blocked 5-fold cross-validation of the closed-form ceiling (contiguous folds, since adjacent token
windows are correlated). The fold-to-fold std is small --- mean 0.024, max 0.062 $R^2$ across all
36 blocks --- and the well-recovered blocks are the tightest (GPT-2 block 2: $0.996 \pm 0.000$),
while the larger spreads sit on the low-recoverability blocks (GPT-2 block 10: $0.434 \pm 0.062$),
as expected. This is the honest companion to the near-zero seed spreads (Section~\ref{sec:setup}):
a genuine data-split CI, an order of magnitude larger than the seed-std but still small enough that
every conclusion stands.

\section{Discussion}
\label{sec:discussion}

\begin{itemize}
\item \textbf{Linear recoverability is heterogeneous, learned, and model-specific --- not set by
  the activation function.} Across 36 blocks of three models the linear ceiling ranges from
  ${\sim}0.2$ to $>0.99$ with no monotone depth trend, and two same-size GELU models (GPT-2,
  Pythia-160m) have opposite profiles. Which FFN blocks reduce to a single linear map is a property
  of the trained network, not of ``GELU vs SwiGLU.'' This is the paper's main empirical message and
  the reason a careful (closed-form) baseline matters.
\item \textbf{The residual is not low-order multiplicative --- and multiplicative form does not
  predict recoverability.} A low-rank bilinear probe recovers only a few points of $R^2$ and its
  gain is uncorrelated with how nonlinear the block is (Pearson $r\approx 0$). An explicitly
  multiplicative SwiGLU block is no more linearly recoverable, and its residual no more bilinearly
  recoverable, than a GELU one. What a single position-wise layer cannot capture is consistent with
  higher-order or distributed computation not a missing single product term. The same holds
  along the \emph{depth} axis:
  a two-layer \emph{multiplicative} candidate (poly ($2\times$)) matches a two-layer \emph{additive}
  one (dense ($2\times$)) to within $\le 0.002$ $R^2$ everywhere (Section~\ref{sec:worked}), so what
  little the 2-layer candidates recover is the extra hidden nonlinearity, not the multiplicativity.
\item \textbf{$R^2$ and downstream $\dppl$ measure different things.} A near-perfect activation fit
  ($R^2$ 0.96) can still cost $+76$ PPL (llama block 0); early blocks are perplexity-critical largely
  independent of fittability, and the multiplicative probe's \emph{downstream} value, where it
  exists, lives on exactly those critical blocks. Studies of FFN compression should report both.
\item \textbf{Closed-form baselines are essential for activation distillation.} Because transformer
  activations are ill-conditioned (outlier features), an under-converged trained linear baseline can
  overstate nonlinearity by tens of points of $R^2$ and an order of magnitude of $\dppl$ --- the
  kind of artifact that can masquerade as strong nonlinearity. The exact least-squares ceiling
  removes the confound and should be standard practice.
\item \textbf{``Recoverable'' comes in two structurally opposite kinds.} A high $\Rlin$ can mean a
  \emph{low-rank, outlier-concentrated} linear block (GPT-2 block 2: $\Rlin$ 0.996 but effective
  rank 1 and per-feature median $R^2$ only 0.16 --- one dominant linear direction, tying it to the
  large-magnitude outlier features of \citealp{dettmers2022llmint8}) or a \emph{high-rank, broadly
  linear} block (Pythia block 0: $\Rlin$ 0.92, effective rank 376, per-feature 0.89). Since $\Rlin$
  and effective rank are decoupled (Section~\ref{sec:kinds}), the variance-weighted recoverability we
  headline should be read together with effective rank and per-feature $R^2$ to know \emph{what kind}
  of linear object a block is. The low-rank blocks resemble single linear key--value memory
  directions \citep{geva2021kv} and compress further still. (The raw weight spectrum alone is
  uninformative here. It is scale-dominated by the outlier feature, so we measure rank by
  reduced-rank regression, not by SVD of $W^{*}$.)
\end{itemize}

\section{Limitations}
\label{sec:limitations}

Base models at the small end (GPT-2, Pythia-160m, llama-160m) and modest corpora (WikiText-2,
Gutenberg prose, and a math/logic puzzle corpus. All English); a TinyLlama-1.1B scale probe but
no large-model survey; perplexity on a capped test slice; healing introduces an in-domain adaptation
confound addressed but not eliminated by the heal-original control. The residual probe is a
\emph{single} low-rank bilinear form. A different basis (higher rank, other nonlinearities) might
recover more of the residual; our negative result is specific to low-order bilinear recovery, which
is the natural first probe. The closed-form baseline is exact only for the \emph{linear} candidate;
trained candidates are verified converged against it but could be optimised further.

\emph{On corpus.} We measured on three domains; WikiText-2 (encyclopedic), Gutenberg prose, and a
math/logic puzzle corpus. The per-block profile is highly stable across all three (Pearson
0.95--0.99, Section~\ref{sec:robust}), so the \textbf{per-block profile is largely stable across the
sampled activation distributions} rather than an artifact of any one corpus. A still-larger or more
varied corpus (WikiText-103, OpenWebText, source code) would test whether the profile holds yet more
broadly, and would tighten the absolute perplexity estimates and reduce the in-domain healing
confound; our domains are so far all English natural-language text. A broader \emph{model} sweep would do more than a larger corpus to chart which
trained networks (and which blocks) admit single-layer linear distillation.

\section{Future Work}
\label{sec:future}

\begin{itemize}
\item A larger model sweep (more architectures and scales) to map how the linear-recoverability
  profile varies and whether ``learned, not architectural'' holds at scale.
\item \textbf{Higher-\emph{degree} single-polynomial layers.} Our poly probe is degree-2 (a sum of
  low-rank bilinears); since Section~\ref{sec:residual} shows the residual is not degree-2-recoverable,
  the natural next probe raises the \emph{degree} within one position-wise layer --- a degree-3+
  factorized polynomial (higher-order factorization machines; \citealp{blondel2016hofm}) rather than
  \emph{stacking} layers (the additive depth of dense ($2\times$) / the multiplicative depth of poly
  ($2\times$), Section~\ref{sec:worked}). This separates whether the residual is higher-order
  \emph{product} structure (recoverable by raising the degree) from genuinely non-polynomial
  computation (recoverable only by an added nonlinearity).
\item \textbf{Mechanism behind the low-rank / broadly-linear split (Section~\ref{sec:kinds}).} We
  \emph{measure} that some recoverable blocks are near-rank-1 (outlier-concentrated) and others
  full-rank; \emph{why} a given block lands in one regime and whether the low-rank blocks
  correspond to identifiable key--value memories \citep{geva2021kv} or to specific token/feature
  roles is open. A weighted/whitened effective rank (RRR is variance-weighted, so it inherits the
  outlier-weighting) would complement the per-feature view.
\item Richer residual bases beyond polynomials (small kernel / feature maps); multi-block span
  distillation (replace several FFNs at once); and a TinyLlama / larger-SwiGLU external-validity check
  with scale-aware fitting.
\item \textbf{A genuine higher-order / geometric-product (Sigma-Pi) layer --- deferred, with tempered
  expectations.} A single log-space weighted product was numerically unstable on these targets and,
  once stabilised, still added no value over the linear ceiling --- unsurprising given
  Section~\ref{sec:residual}: a \emph{single monomial} is an even more restrictive form than the
  sum-of-bilinears that already fails to recover the residual, so we do not expect a stable
  reformulation to help on FFN residuals specifically. Its likely value lies elsewhere; on targets
  with genuine low-order \emph{product} structure (bilinear attention scores, explicit gating, or the
  weight-generation hypernetwork of our concurrent work \citep{whipp2026pi}, where the recruitment
  gate \emph{did} fire) rather than on the FFN residual studied here.
\end{itemize}

\section{Reproducibility}
\label{sec:repro}

All experiments are implemented in the \texttt{polyweave} codebase, which will be released with the paper. Per-block fitting is
\texttt{polyweave/experiments/gpt2\_mlp\_distill.py}; the depth survey
(Sections~\ref{sec:survey}--\ref{sec:residual}, Figure~\ref{fig:depth}) is
\texttt{run\_depth\_sweep.py} (re-plot without recompute via \texttt{plot\_depth\_sweep.py}); the
residual-gain probe (Section~\ref{sec:residual}, Figure~\ref{fig:residual}) is
\texttt{run\_residual\_gain\_clean.py} (frozen-linear, quad-only, early-stopped); the effective-rank /
per-feature analysis (Section~\ref{sec:kinds}, Figure~\ref{fig:kinds}) is \texttt{run\_rrr\_rank.py}
(closed-form + reduced-rank regression, no training; re-plot via \texttt{plot\_rrr\_rank.py}); the
cross-domain robustness (Section~\ref{sec:robust}, Figure~\ref{fig:corpus}) is
\texttt{run\_corpus\_robustness.py} (closed-form ceilings on Moby-Dick / math-puzzle corpora; overlay
via \texttt{plot\_corpus\_robustness.py}), \texttt{run\_depth\_sweep\_gutenberg.py} (full poly +
$\dppl$ sweep on Moby-Dick with its own train/test split), and \texttt{run\_kfold\_ceilings.py}
(blocked 5-fold data-split CI); the two-block detail and perplexity
(Sections~\ref{sec:worked}--\ref{sec:ppl}) are driven multi-seed by
\texttt{run\_gpt2\_multiseed\_v2.py}, \texttt{run\_pythia\_multiseed\_v2.py}, and
\texttt{run\_llama\_multiseed\_v2.py}; the multiplicative-vs-additive depth control
(Section~\ref{sec:worked}, poly ($2\times$)) is \texttt{run\_poly2x.py}; the TinyLlama-1.1B scale
probe (Section~\ref{sec:worked}) is \texttt{run\_tinyllama\_scaleaware.py}. The linear baseline is the
exact closed-form least-squares solution (\texttt{linear\_closed\_form=True}); trained candidates use
8\,000 AdamW steps. Optional deps install via \texttt{pip install polyweave[distill]}
(\texttt{transformers}, \texttt{datasets}); WikiText-2 is cached to plain text on first fetch. Runs
target a single 6\,GB GPU.

\bibliographystyle{plainnat}
\bibliography{references_paper2}

\end{document}